\documentclass{article}

\usepackage{PRIMEarxiv}

\usepackage[utf8]{inputenc} 
\usepackage[T1]{fontenc}    
\usepackage{hyperref}       
\usepackage{url}            
\usepackage{booktabs}       
\usepackage{amsfonts}       
\usepackage{nicefrac}       
\usepackage{microtype}      
\usepackage{lipsum}
\usepackage{fancyhdr}       
\usepackage{graphicx}       
\graphicspath{{media/}}     
\usepackage{caption}
\usepackage{subcaption}
\usepackage{algorithm2e}
\RestyleAlgo{ruled}
\usepackage{graphics}
\usepackage{makecell}
\usepackage{stix}
\usepackage{multirow}

\usepackage[dvipsnames]{xcolor}

\newsavebox\CBox

\title{MIAS-SAM: Medical Image Anomaly Segmentation without thresholding}

\author{
  Marco Colussi~\textsuperscript{1} ~\thanks{Corresponding author: \texttt{marco.colussi@unimi.it}.}
  \And
  Dragan Ahmetovic\textsuperscript{1} \\ \\
  \textsuperscript{1} Università degli studi di Milano\\
  \And
  Sergio Mascetti\textsuperscript{1} \\
}

\begin{document}
\maketitle

\begin{abstract}
This paper presents MIAS-SAM, a novel approach for the segmentation of anomalous regions in medical images.
MIAS-SAM uses a patch-based memory bank to store relevant image features, which are extracted from normal data using the SAM encoder.
At inference time, the embedding patches extracted from the SAM encoder are compared with those in the memory bank to obtain the anomaly map.
Finally, MIAS-SAM computes the center of gravity of the anomaly map to prompt the SAM decoder, obtaining an accurate segmentation from the previously extracted features.
Differently from prior works, MIAS-SAM does not require to define a threshold value to obtain the segmentation from the anomaly map.
Experimental results conducted on three publicly available datasets, each with a different imaging modality (Brain MRI, Liver CT, and Retina OCT) show accurate anomaly segmentation capabilities measured using DICE score.
The code is available at: \url{https://github.com/warpcut/MIAS-SAM}
\end{abstract}

\keywords{Anomaly segmentation \and Medical Imaging \and Segment anything \and Anomaly detection}

\section{Introduction}
\label{sec:intro}

In recent years, there has been growing interest in the application of anomaly detection techniques for the specific purpose of segmenting anomalous regions~\cite{cao2023segment,Ma_2025_WACV}.
This approach is particularly relevant in the medical domain since manual annotation of anomalous regions requires expert domain knowledge, and is also time-consuming which makes it impractical for large-scale datasets.
Indeed, traditional supervised segmentation approaches require medical practitioners to annotate pixel-level anomaly masks.
Instead unsupervised anomaly segmentation methods automatically identify potential anomalous regions in the images by learning from non-anomalous samples only~\cite{baur2021autoencoders,bercea2024diffusion}.


Existing methods that employ anomaly detection techniques for segmentation typically compute the anomalous region by thresholding the anomaly map~\cite{bercea2024denoising,bao2024bmad,Ma_2025_WACV}.
%
In such cases, the threshold should be ideally selected using a validation set to ensure consistency and accuracy. However, in unsupervised anomaly detection settings, such a validation set is usually not available. As a consequence, thresholds are often chosen arbitrarily (\textit{e.g.}, a fixed value of $0.5$), and this can result in suboptimal choices leading to imprecise segmentation, particularly around object borders, where anomalies may blend into normal regions~\cite{bao2024bmad}.
An alternative solution is to use post-hoc optimization, such as selecting the threshold value that maximizes the DICE metric on test data, thus representing an upper bound for the results but possibly not the actual performance that could be obtained in real-world applications~\cite{behrendt2024leveraging,bercea2024diffusion,colussi2024loris,Ma_2025_WACV}.

To address the above problems, this paper presents MIAS-SAM (Medical Image Anomaly Segmentation with SAM), a feature-level anomaly segmentation technique that does not rely on a threshold value to compute the anomaly segmentation.
For anomaly map generation, MIAS-SAM uses a memory bank to store patch embeddings~\cite{roth2022towards} extracted from a ViT encoder~\cite{dosovitskiy2020image} from the train data, and subsequently compares them with the ones obtained from test data.
To compute the segmentation, MIAS-SAM computes the anomaly map center of gravity and uses it, together with the previously extracted embeddings, as a point prompt for the SAM decoder.

MIAS-SAM has three main advantages over existing techniques.
First, unlike other SAM-based anomaly segmentation methods that use multimodal text-vison models~\cite{li2025clipsam},
MIAS-SAM automatically computes the point prompt, and therefore does not require a textual prompt.
Using a textual prompt would require prompt-engineering, a procedure that would need to be be repeated for each dataset~\cite{Li_2024_BMVC}.
Second, unlike many previous techniques, MIAS-SAM does not rely on a threshold value to compute the anomaly segmentation from the anomaly map.
Third, MIAS-SAM produces precise segmentations, as demonstrated by the experimental results conducted on three public datasets.

Our contribution can be summarized as follows.

\begin{itemize}
    \item We present a solution to extend patch based anomaly detection techniques to use SAM ViT encoder.
    \item We propose using the center of gravity of the anomaly map as prompt, removing the need for a segmentation threshold and for a textual prompt.
    \item We evaluate our method on three different benchmark datasets, each for a different imaging modality (Brain MRI, Liver CT and Retina OCT). We show that, against SOTA approaches, MIAS-SAM improves the segmentation accuracy in two of the three datasets.
\end{itemize}

\section{Related work}

Unsupervised anomaly detection approaches are divided into three main categories:
reconstruction-based, synthesizing-based and embeddings-based.
Reconstruction-based methods rely on models trained to reconstruct normal data, detecting anomalies when reconstruction error is high~\cite{baur2021autoencoders}.
Synthesizing-based methods generate synthetic anomalies on normal images to model the anomalous behavior of the data and train a disciminator to differentiate between the two~\cite{li2021cutpaste}.
Embedding-based methods learn feature representations, distinguishing normal from anomalous patterns within the latent space. These approaches have recently garnered significant attention, particularly due to their ability to leverage pre-trained models for extracting generalizable features, as anomalies may reside in the feature space~\cite{liu2023simplenet,guo2024recontrast}. For this reason, the technique proposed in this paper adopts this approach.

\paragraph{Embedding based.} 
An example of embedding-based technique that we extend in this work is PatchCore~\cite{roth2022towards} that generates patches of normal features and creates a memory bank that characterizes normal patterns.
Then, it measures the similarity between the patched features extracted from an image in the test-set with the stored ones, using similarity as the anomaly measure.
In this paper we compare the performance of our proposed solution with the following embedding-based techniques. CFA~\cite{lee2022cfa} facilitates transfer learning by adapting patch representations to the target dataset and then creates a memory bank to compute the patches similarity.
SimpleNet~\cite{liu2023simplenet} combines synthetic and embedding-based method by generating anomalous features adding noise to the embedding extracted from a frozen pre-trained encoder and training a discriminator to identify the anomalous features.
RD4AD~\cite{deng2022anomaly} and ReContrast~\cite{guo2024recontrast} focus on reconstructing the features extracted from a pre-trained encoder, and using the similarity of the reconstructions as anomaly scores.
Finally, CFlow-AD~\cite{gudovskiy2022cflow} learns to convert the normal feature distribution into a Gaussian distribution using normalizing flows, then anomaly scores are assigned based on their distance from the distribution mean.

\paragraph{Foundational models.}
A recent approach to anomaly detection for segmentation leverages promptable foundational vision models such as SAM~\cite{kirillov2023segment} and multimodal text-vision models such as CLIP~\cite{radford2021learning}.
This approach focuses on the models' zero-shot capabilities and their ability to produce robust and transferable feature representations.
Such approaches rely on the ability of the text encoder to extract embeddings that are aligned with the image features.
The problem is that these text encoders are strictly dependent from the textual prompt that thus needs to be carefully engineered.
This can be achieved either via specific knowledge of the target domain~\cite{cao2023segment}, complex prompt templates~\cite{jeong2023winclip}, or prompt learning~\cite{li2024promptad,cao2024adaclip}.
These contributions partially address the problem of prompt engineering, and at least a part of the textual prompt needs to be manually defined for each type of data (\textit{e.g.}, ``bottle with defect'', ``bottle with crack'').
Zero-shot anomaly detection has also been applied in the medical domain with
PPAD~\cite{sun2024position} that relies on CLIP and adds learnable text and image prompts to bridge the gap between pre-training data and medical image data.
Another technique is ClipSAM~\cite{li2025clipsam} which first extracts anomaly proposals using CLIP, and then applies a threshold value to the anomaly map to extract a bounding box, which is finally used to prompt SAM thus obtaining the anomaly segmentation.

Our approach is also based on SAM, but differs from other SAM-based methods such as ClipSAM~\cite{li2025clipsam} and SAA+~\cite{cao2023segment}, in the way it generates segmentation prompts.
Specifically, our approach does not rely on textual prompt engineering which is a central component in all the aforementioned techniques.
Textual prompt engineering presents various constraints~\cite{Li_2024_BMVC}: 1) its effectiveness is strictly related to the quality of the specific prompts, 2) it requires involvement of domain experts for the specific data, and 3) the procedure should be repeated for each dataset.
For these reasons, we do not experimentally compare MIAS-SAM with the solutions based on textual-prompt.

Another relevant difference between MIAS-SAM and the anomaly segmentation techniques described in this section is that MIAS-SAM does not require to define a threshold value to compute the anomaly segmentation from the anomaly map.

\section{Method: MIAS-SAM}

\subsection{Problem formulation}
In this work we address the problem of unsupervised anomaly segmentation in which $X_{train}$ is a set of normal images forming the training set and $D=\langle X_{test}, Y, Z \rangle$ is the test set, where $X_{test}$ is a set of normal or anomalous images, $Y : X_{test} \rightarrow [0,1]$ is a function representing the ground-truth annotations such that, for each image $x \in X_{test}$, $Y(x) = 0$ if $x$ is normal and $Y(x) = 1$ if $x$ is anomalous.
Finally, $Z$ is a function that, for each image $x \in X_{test}$, returns the ground truth segmentation mask, that is, a binary image having the same dimensions as $x$. Each pixel $p$ of $Z(x)$ has value $1$ if the corresponding pixel in $x$ is anomalous, $0$ otherwise.
Note that, given a normal image $x \in X_{test}$, all pixels in $Z(x)$ have value zero.
The objective, given an image $x \in X_{test}$, is to predict, for each pixel $p \in x$, whether $p$ is anomalous.

\begin{figure}[ht]
\centering
\includegraphics[width=\textwidth]{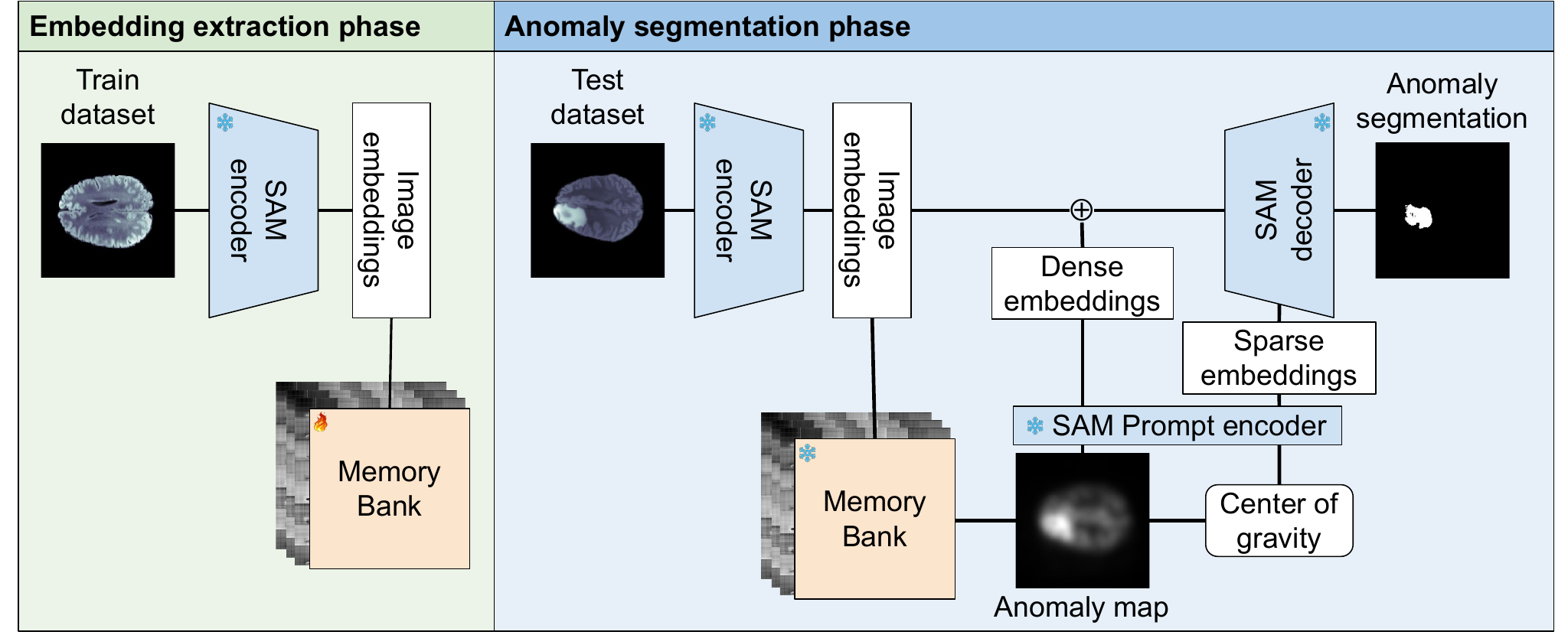}
\caption{\textbf{Overview of the MIAS-SAM methodology}. First, image embeddings are extracted using the SAM encoder and patches are extracted and stored in a memory bank. At test time, images are encoded, and anomaly map are generated based on patch-wise distances. The anomaly maps are then used to compute a spatial prompts, which guide the decoder in segmenting the anomalous region.}
\label{fig:method}
\end{figure}

Fig.~\ref{fig:method} shows the overall architecture of our method. It is composed of two main phases (embedding extraction and anomaly segmentation) and four main components (the SAM encoder $f_\theta$, a memory bank $\mathbf{M}$, the SAM prompt encoder and the SAM decoder).

\subsection{Embedding extraction phase} 

In the \textbf{embedding extraction phase}, the SAM encoder takes in input an image and generates its embeddings $\mathbf{e}_{img}$.
Unlike PatchCore~\cite{roth2022towards}, which uses a ResNet-based encoder and requires multi-layer features to preserve both high-level semantics and fine details, MIAS-SAM extracts features solely from the last layer of the ViT encoder that is pre-trained with masked autoencoders on a large dataset~\cite{kirillov2023segment}. This simplifies the architecture while maintaining both local and global context, thanks to the ViT's attention-based modeling~\cite{dosovitskiy2020image}. The extracted embeddings are of shape $256 \times 64 \times 64$.

The embeddings are then processed with the patching function $\mathbf{P}$ that generates patches with size $5 \times 5$ and a stride of $2$. The patches are stored in the memory bank and indexed with Faiss~\cite{douze2024faiss}.
At the end of the training process, the memory bank contains all the patches obtained from the embeddings of all images in the training set.

\begin{equation}
    \mathbf{M}=\bigcup_{i=0}^{X_{train}}\mathbf{P}(f_\theta(x_i))
\end{equation}

\subsection{Anomaly segmentation phase}
\label{sec:anomaly_seg}
In the \textbf{anomaly segmentation phase} each image in the test set is processed by the same encoder used in the embedding extraction phase. 
The resulting embeddings are then patched with $\mathbf{P}$.
Similarly to PatchCore~\cite{roth2022towards}, each resulting patch $p$ is assigned an anomaly score by computing the maximum distance of $p$ from its nearest neighbor patches in the memory bank $\mathbf{M}$.
The anomaly map is then generated by interpolating each patch to its original location.

Starting from the anomaly map, the SAM prompt encoder generates dense and sparse embeddings.
The dense embeddings are obtained directly from the anomaly map.
Instead, to generate the sparse embeddings, MIAS-SAM first computes the weighted center of gravity $\langle x_c, y_c \rangle$ of the anomaly map as follows.

\begin{equation}\label{eq:cog}
    x_c = \frac{\sum_{i=1}^{H} \sum_{j=1}^{W} j \cdot I(i,j)^\gamma}{\sum_{i=1}^{H} \sum_{j=1}^{W} I(i,j)^\gamma}, y_c = \frac{\sum_{i=1}^{H} \sum_{j=1}^{W} i \cdot I(i,j)^\gamma}{\sum_{i=1}^{H} \sum_{j=1}^{W} I(i,j)^\gamma}
\end{equation}
where $H$ and $W$ are the height and width of the anomaly map, respectively, $I(i, j)$ represents the intensity value in the anomaly map at pixel $\langle i, j \rangle$, and $\gamma$ is a weighting factor that emphasizes higher-intensity pixels, whose value was set to $5$ in the experiments reducing noise from background pixels.
Then, the SAM prompt encoder extracts the positional embeddings from the center of gravity.

To generate the anomaly segmentation MIAS-SAM first sums the dense embeddings with the image embeddings previously computed by the encoder. The resulting embeddings are fed to the SAM decoder together with the sparse embeddings. 
The decoder then generates three different predictions that represent three levels of depth in the proposed segmentation, from the coarser (whole) to the finest (sub-part) \cite{kirillov2023segment}.
Among the three hierarchical outputs from the SAM decoder, we empirically found that the third mask aligns with the finest granularity of anomalies (see Section~\ref{sec:ablation_mask}).

\section{Experimental evaluation}

We compared MIAS-SAM with six state-of-the-art embedding-based anomaly detection approaches, using th methodology is described in Section~\ref{sec:eval}.
The evaluation was performed on three benchmark datasets (described in Section~\ref{sec:dataset}), each based on a different medical imaging modality.
The results, described in Section~\ref{sec:res}, compare the models' ability to identify the anomaly, and to compute the anomaly map and the anomaly segmentation.

\subsection{Evaluation methodology}
\label{sec:eval}

\paragraph{Implementation details.} We use the Segment Anything Model (SAM)~\cite{kirillov2023segment} with the ViT-B~\cite{dosovitskiy2020image} backbone as our base segmentation model, which was pretrained on a large and diverse dataset of over 11 million natural images with more than 1 billion segmentation masks. The images are first rescaled to a fixed resolution of $1024x1024$ to fit the expected input of SAM and then normalized between [0,1]. Experiments were run on an NVIDIA GPU A100 with 80GB memory. The system is implemented in Python $3.10.12$ using PyTorch $2.0.1$.

\paragraph{Baselines.}
We compared the performance of MIAS-SAM with 6 state of the art embeddings-based anomaly detection techniques: RD4AD~\cite{deng2022anomaly}, PatchCore~\cite{roth2022towards}, CFA~\cite{lee2022cfa}, CFlow-AD~\cite{gudovskiy2022cflow}, SimpleNet~\cite{liu2023simplenet} and ReContrast~\cite{guo2024recontrast}.
For the first five techniques we use the results reported in the benchmark comparison by Bao et al.~\cite{bao2024bmad}.
Note that the benchmark comparison uses a threshold of $0.5$ on the anomaly map to obtain the anomaly segmentation results.
For ReContrast~\cite{guo2024recontrast} (which is not included in the benchmark comparison), we run the technique on the selected datasets. Also in this case, we adopt a fixed threshold of $0.5$ on the anomaly map.
This ensures a consistent and unbiased comparison across methods.

\paragraph{Evaluation metrics.}
We used the following metrics to evaluate the various methods:
\begin{itemize}
    \item Pixel-level AUROC (P-AUROC): measures the model ability to localize anomalies by distinguishing anomalous and normal pixels in the anomaly map.
    \item DICE score: evaluates anomaly segmentation accuracy by measuring the overlap between predicted and ground truth anomaly regions.
\end{itemize}

\subsection{Datasets}
\label{sec:dataset}

The datasets were made available by Bao et al~\cite{bao2024bmad} and they are specifically designed for the anomaly detection task in the medical domain.
Among the six available datasets, we selected those having ground truth anomaly segmentation masks.
Each dataset contains a training set of non-anomalous images, and a test set with both anomalous and non-anomalous images.
The datasets are based on imaging techniques that generate 3D representations of the body part that are then sliced to generate 2D images.

The \textbf{Brain MRI} dataset is based on the BraTS2021 dataset~\cite{baid2021rsna}.
Magnetic resonance imaging (MRI) uses strong magnetic fields and radio waves to form 3D volumes of organs in the body. 
The training set is composed of $7,500$ slices of normal brain MRIs while the test set is composed of $3715$ slices of brains with tumors, along with their segmentation masks. Note that the anomalies are visible in a subset of the slices ($3075$).

The \textbf{Liver CT} dataset is obtained by combining BTCV~\cite{bilic2023liver} and LiTs~\cite{landman2015miccai} datasets and extracting liver regions only.
Computed tomography (CT) uses X-rays and computational processing to generate the body's internal structures.
The former dataset (BTCV~\cite{bilic2023liver}) does not contain anomalies, hence its $1,542$ slices were used for training, while $1,493$ slices of the latter dataset (LiTs~\cite{landman2015miccai}) were used for testing, $660$ of which contain an anomaly.

The \textbf{Retina OCT} dataset was generated from the RESC dataset~\cite{hu2019automated} of OCT images.
Optical Coherence Tomography (OCT) is a non-invasive technique that uses low-coherence light to generate high-resolution images of biological tissues.
The dataset contains both normal images and others in which retinal edema is visible. The training set is composed of $4,297$ normal images while the test set is composed of $764$ anomalous samples and $1805$ normal ones.

\subsection{Results}
\label{sec:res}

\begin{table}[]
\begin{center}
\resizebox{\textwidth}{!}{%
\begin{tabular}{l|cc|cc|cc}
\toprule
\multirow{2}{*}{Method} & \multicolumn{2}{c|}{Brain MRI} & \multicolumn{2}{c|}{Liver CT} & \multicolumn{2}{c}{Retina OCT} \\
\cline{2-7}  & P-AUROC & DICE & P-AUROC & DICE & P-AUROC  & DICE \\
\midrule
RD4AD (CVPR22) & \underline{96.45} & 28.28 & 96.01 & 10.72  & 96.18 & 33.51 \\
PatchCore (CVPR22) & \textbf{96.97} & \underline{32.82} & 96.43 & 10.49 & 96.48 & \textbf{57.04} \\
CFA (Access22) & 96.33 & 30.22 & 97.24 & 14.93 & 91.10 & 36.57 \\
CFlow-AD (WACV22) & 93.76 & 19.50 & 92.41 & 7.58 & 93.78 & 44.83 \\
SimpleNet (CVPR23) & 94.76 & 28.96 & 97.51 & 12.26 & 77.14 & 30.28 \\
ReContrast (NiPS23) & 95.64 & 12.57 & \textbf{98.02} & \underline{15.03} & \underline{97.96} & 25.02 \\
\midrule
MIAS-SAM (Ours) & 96.35 & \textbf{37.04}  \textcolor{green}{$\blacktriangle\sim13\%$} &\underline{97.56} & \textbf{42.85} \textcolor{green}{$\blacktriangle\sim185\%$}& \textbf{99.76} & \underline{51.21} \textcolor{red}{$\blacktriangledown\sim11\%$}\\

\bottomrule
\end{tabular}
}
\end{center}
\caption{Anomaly segmentation performance of MIAS-SAM in comparison with baselines on brain, liver and retina datasets. Best results are reported in \textbf{bold}, second-best are \underline{underlined}.
}
\label{tab:comparison}
\end{table}

Table~\ref{tab:comparison} shows results of the quantitative comparison.
Considering the P-AUROC, there is not a single technique that outperforms the others in all three datasets.
Specifically, MIAS-SAM is the best-performing model on the Retina OCT dataset, obtaining a score of $99.76$ and an improvement of $1.84$\% over the second best model (ReContrast).
For the Liver CT dataset MIAS-SAM had the second best result, with a score of $97.56$, losing $0.47$\% with respect to ReContrast.
For the Brain MRI dataset, MIAS-SAM had the third best result, with a score of $96.35$ and a loss of $0.64$\% with respect to PatchCore.
We also highlight that in all the datasets the variability in the scores is limited (stdev = [$1.04 : 7.05$]).
This confirms the previous findings that question the ability of the P-AUROC metric to robustly assess the quality of the segmentation techniques, suggesting to favour the DICE score metric instead~\cite{Ma_2025_WACV}.

Considering the DICE score metric,
MIAS-SAM performs as the best technique in two datasets and as the second-best in the third one.
In particular, MIAS-SAM shows a gain of $4.2$ points ($+13\%$) with respect to the second-best in the Brain MRI dataset, and an increase of $27.8$ points ($+185\%$) on the Liver CT dataset.
Instead, in Retina OCT datasets MIAS-SAM is second-best performing model with a decrease of $5.8$ points ($-11\%$) with respect to PatchCore, which is the best-performing.

\begin{figure}[ht]
\centering
\includegraphics[width=\textwidth]{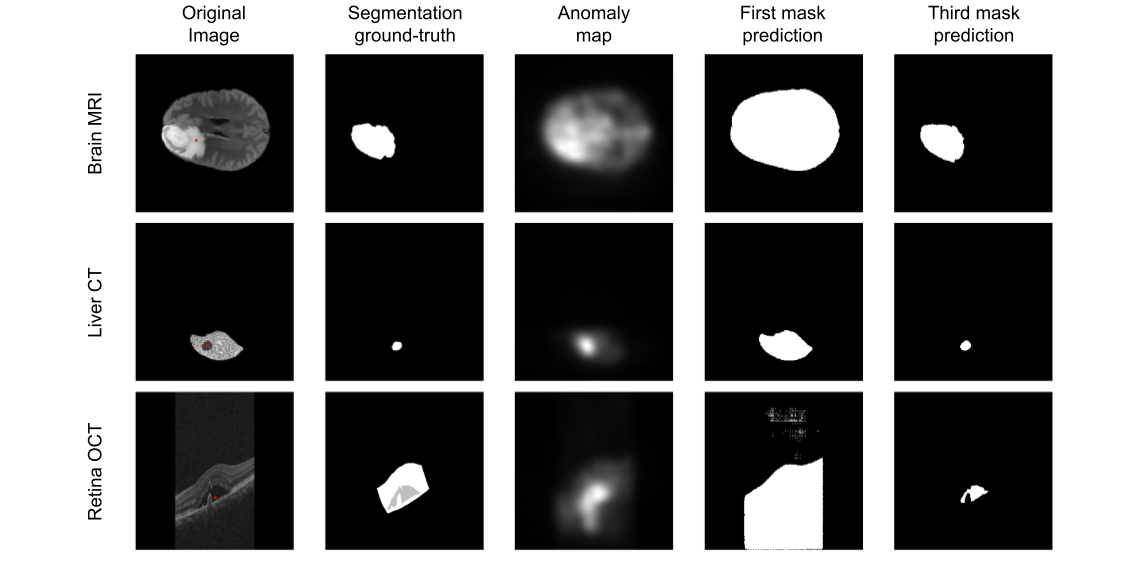}
\caption{\textbf{Qualitative results of MIAS-SAM.} Each row corresponds to a different dataset: brain (top), liver (middle), and retina (bottom). The first column shows the input image, the second the generated anomaly map, the third the first segmentation mask, and the fourth the third segmentation mask, that best captures the anomaly.}
\label{fig:comparison}
\end{figure}

Fig.~\ref{fig:comparison} shows three examples from each dataset. Each example includes the original image with the computed center of gravity (the red dot), the ground truth anomaly segmentation mask, the anomaly map computed by MIAS-SAM, and the predictions of the first and third masks returned by the SAM decoder.
The computed center of gravity is represented as a red point in the original image.
Considering the first two rows (Brain MRI and Liver CT) we observe that, in the considered examples, the center of mass is within the ground truth anomaly segmentation and that the third mask (last column in the image) correctly segments the anomaly.
The example is the last row can provide insights on why MIAS-SAM underperforms in the Retina OCT dataset.
First,
we can observe that the ground truth reports two segmented areas, where the white represents the anomaly area and the gray one is actual retinal edema.
In accordance with the work of Bao et al.~\cite{bao2024bmad}, we considered the white area as the ground truth (note, the white area is reported in all images, while the gray area is reported in some images only). However, in some cases (as the one shown in Fig.~\ref{fig:comparison}) the segmentation provided by MIAS-SAM extracts only the presence of liquid inside the retina, represented in gray in the ground truth.
Second, there is no visual boundary differentiating the anomalous (white) region from the non-anomalous one, thus making it difficult for the SAM decoder to identify the target area.

\subsection{Ablation studies}
\paragraph{Effect of Mask Selection on Segmentation Performance.} During the experimental evaluation we also compared the performance on MIAS-SAM with the three masks returned by the SAM decoder.
Table~\ref{tab:mask} shows that our proposed technique always performs better with the last maks.
As described in the original SAM paper~\cite{kirillov2023segment}, each mask represents a different level of depth in the object. We noticed that using point prompt the first mask often represents the whole anatomical area (\textit{e.g.}, the brain segmentation from the background), the second one a more specific area of the image (\textit{e.g.}, half the brain), and the third the actual anomaly (\textit{e.g.}, brain tumor).

\begin{table}[]
\begin{center}
\setlength{\tabcolsep}{6pt}
\begin{tabular}{c|ccc}
\toprule
Mask \# & Brain MRI & Liver CT & Retina OCT \\
\midrule
1    & $23.61$ & $15.05$ & $26.08$ \\ \hline
2    & $32.65$ & $25.32$ & $48.06$\\ \hline
3    & $\textbf{37.06}$ & $\textbf{42.85}$ & $\textbf{51.21}$ \\
\bottomrule
\end{tabular}
\end{center}
\caption{Results for different mask generated by SAM decoder (DICE).}
\label{tab:mask}
\end{table}

\paragraph{Effect of prompt.}\label{sec:ablation_mask}
We compare MIAS-SAM with two alternative ways for computing the encoder prompt: 1) the point with maximum anomaly value and 2) the bounding box of the anomalous region.
With the maximum anomaly approach, the prompt consist of a single point that has the maximum anomaly value in the anomaly map. 
In the bounding box approach, 
the anomaly map is first binarized with a fixed threshold of $0.5$ and then the bounding box of the resulting area is used as the prompt.
If multiple disconnected regions are identified, a separate bounding box is computed for each blob and individually provided as input to the prompt encoder.

Table \ref{tab:prompt} shows the results for the ablation study comparing how the three ways of computing the prompt impact the DICE score (note that the P-AUROC does not change, as it is computed on the anomaly map).
The solution based on the center of gravity outperforms the bounding box approach on the Brain and Liver datasets. A possible motivation is that the fixed threshold used in the bounding box approach can be sub-optimal, as discussed in Section~\ref{sec:intro}.
In contrast, using the bounding box yields better results on the Retina dataset, thus supporting the motivation provided in Section~\ref{sec:res}. 
Interestingly, for this dataset, the bounding box approach outperforms the baselines.
Table \ref{tab:prompt} also shows that using the maximum anomaly as prompt result in worse performance. A possible motivation is that the highest-scoring pixel often lies at the boundary of the anomalous region, which can reduce the model's ability to accurately determine whether the prompt corresponds to the anomaly itself or the surrounding context.
In contrast, using the center of gravity provides a more reliable point prompt, as it is consistently located within the core of the anomalous region. This central positioning offers a more representative and unambiguous cue to the model, leading to improved segmentation performance.

\begin{table}[]
\begin{center}
\setlength{\tabcolsep}{6pt}
\begin{tabular}{c|c|ccc}
\toprule
Method & Prompt type & Brain MRI & Liver CT & Retina OCT \\
\midrule
Thresholding & Bounding box & $15.93$ & $14.81$ & $\textbf{65.80}$ \\ \hline
Maximum anomaly & Point & $31.54$ & $42.70$ & $38.11$\\ \hline
Center of Gravity (ours) & Point & $\textbf{37.06}$ & $\textbf{42.85}$ & $51.21$ \\
\bottomrule
\end{tabular}
\end{center}
\caption{DICE score results for different ways of generating the SAM prompt.}
\label{tab:prompt}
\end{table}

\section{Conclusion}
This paper presents Medical Image Anomaly Segmentation with SAM (MIAS-SAM), which is an embeddings-based anomaly segmentation technique.
MIAS-SAM adopts an innovative pipeline, which includes a novel prompt generation approach, to extract accurate anomaly segmentations through SAM.
MIAS-SAM does not require to threshold the anomaly map to segment the images,
hence addressing a common problem in many existing anomaly detection techniques.
Furthermore, it does not require to perform dataset-specific textual prompt-engineering which is the current standard when using foundational models for the anomaly segmentation task.

The experiments,
conducted on three benchmark datasets, each with a different imaging modality (Brain MRI, Liver CT and Retina OCT),
show that MIAS-SAM
outperforms other techniques on two datasets and achieves state of the art performance on the third one.

Our approach is the first to generate anomaly maps from SAM encoder embeddings and to use them as point-prompts for guiding the anomaly segmentation.
The increased performance demonstrates the potential of this strategy.
Future work will further aim at improving the approach used to obtain the anomaly map, with the goal of obtaining better prompts and improving the resulting segmentation accuracy.

Another possible improvement would be to use foundation models specifically designed for medical images.
We experimented with MedSAM~\cite{ma2024segment} that, however, resulted in lower performance.
Another possible improvement is to use negative point prompts to indicate which areas of the image should be considered as background, thus possibly increasing the quality of the segmentations.

\section*{Acknowledgement}
This project was partially supported by TEMPO – Tight control of treatment efficacy with tElemedicine for an improved Management of Patients with hemOphilia project, funded by the Italian Ministry of University and Research, Progetti di Ricerca di Rilevante Interesse Nazionale (PRIN) Bando 2022 - grant [2022PKTW2B].

\bibliographystyle{unsrt}  
\bibliography{paper}  

\end{document}